\documentclass[twocolumn]{webofc}

\usepackage[varg]{txfonts}   
\usepackage{hyperref}
\usepackage{url}
\hypersetup{colorlinks=true,citecolor=blue,urlcolor=blue,linkcolor=blue}
\usepackage{amsmath}
\usepackage{amssymb}
\usepackage{commath}
\usepackage[ruled,norelsize,vlined,linesnumbered]{algorithm2e}
\makeatletter
\newcommand{\removelatexerror}{\let\@latex@error\@gobble}

\makeatother
\graphicspath{ {Figures/} }
\begin{document}
\title{Adaptive Twisting Sliding Control for Integrated Attack  UAV's \\Autopilot and Guidance}
%
%

\author{\firstname{Minh Tu} \lastname{Nguyen}\inst{1}\fnsep\thanks{\email{minhtu1709@gmail.com}} \and
		\firstname{Van Truong} \lastname{Hoang}\inst{1}\fnsep\thanks{\email{vantruong.hoang@alumni.uts.edu.au}} \and
        \firstname{Manh Duong} \lastname{Phung}\inst{2}\fnsep\thanks{\email{duong.phung@fulbright.edu.vn}} \and
        \firstname{Van Hoa} \lastname{Doan}\inst{3}\fnsep\thanks{\email{doanvanhoa@gmail.com}}
}

\institute{Faculty of Missile and Gunship, Naval Academy, Nha Trang, Khanh Hoa, Vietnam 
\and
           Undergraduate Faculty, Fulbright University, Ho Chi Minh City, Vietnam 
\and
           Information Technology Institute, Academy of Military Science and Technology, Ha Noi, Viet Nam
          }

\abstract{This paper investigates an adaptive sliding-mode control for an integrated UAV autopilot and guidance system. First, a two-dimensional mathematical model of the system is derived by considering the incorporated lateral dynamics and relative kinematics of the UAV and its potential target of attack. Then, a sliding surface is derived utilizing the zero-effort miss distance. An adaptive twisting sliding mode (ATSMC) algorithm is applied to the integrated system. Simulation and comparisons have been accomplished. The results show our proposed design performs well in interception precision, even with high nonlinearity, uncertainties, disturbances, and abrupt changes in the target's movement, thanks to the adaptation strategy.
}
\maketitle
\section{Introduction}\label{intro}
Attack UAVs, a type of combat drone, are flying objects that are capable of striking one or a number of targets. The current usage and application of this type of UAV have made it of great interest in the literature. The growth of attack UAVs likewise developed in various fields, mechatronics and electronics, control and guidance, data mining and communication, \cite{santoso2020} to name a few. Among these researches, flight control and guidance loops of attack UAVs are particularly important. 

Initially, control and guidance are independently from each other, or they are spectrally separated into an inner-loop autopilot and the outer-loop guidance. However, separation is not able to be proper as it causes instability and miss distances between the UAV and its target due to the instantaneous variation in their relative geometry. In the end, the integrated problem has been continuously investigated as it can improve  the interceptor's performance by solving issues related to high nonlinearirity coupling effect between the guidance and control dynamics. Furthermore, the integrated guidance and autopilot controller (IGAC) can minimize the final miss distance and control energy in the worst conditions related to the unpredictable maneuvers of the target \cite{santoso2020}.

In the last two decades, many researchers have claimed their findings in state-of-the-art IGAC algorithms followed by a number of algorithms and evaluations, such as back-stepping, model predictive, sliding mode, and nature-inspired methods. Back-stepping-based techniques \cite{chao2019, shen2024} are capable robust, and highly accurate, but they may be relevant to static targets only. Model predictive algorithms were proposed to improve properties of maneuverable performance and system uncertainties but they do not guarantee real-time operation due to a weakness in processing time and optimization process. Natural inspiration and learning algorithms  \cite{ma2023deep, Moon2021, yuan2023} can provide opportunities for more powerful and robust IGAC methods, thanks to the complementary nature of knowledge-based and learning power. However, data feeding for these techniques is not usually available, particularly in combat situations. In the end, Sliding mode control (SMC)-based approaches \cite{cross2020, yang2018new, ali2016lateral} are overwhelming in number because they are exemplary solutions to deal with unpredictable movements of the target in a faster convergence. However, by applying sliding mode, the IGAC system may suffer from heightened chattering with uncontrollable frequencies and magnitudes \cite{santoso2020}.

Higher-order sliding modes (HOSM) have been introduced to minimize chattering as well as to enhance the finite-time convergence \cite{Levant1993}. One of the most favored HOSM is the twisting-based controllers \cite{ Polyakov2009, Polyakov2009R, Shtessel2012, Dvir2015}. The HOSM has been devoted to UAV control \cite{Zheng2014, Hoang20172nd}. However, these approaches are usually intricate and difficult to apply in real-world applications. The adaptive twisting, inspired by the accelerated twisting sliding mode (ATSM) was suggested for UAV thanks to its control gain being adjusted to reduce the gain amplitude and it can stimulate convergence in finite-time \cite{Hoang2020}. Inspired by our previous work in that respect, we present in this research an application of the adaptive twisting sliding mode control for the integrated guidance and control of the attack UAV reaching its target in vertical planar. Besides that, the well-known zero-effort miss distance, which does not require any control activity to ensure interception, is applied to control the elevator of the UAV in severe circumstances such as nonlinearity, superficial disturbances, and decisive coupling. The proposed method is demonstrated in simulations and comparisons to confirm its achievement.

The paper is organized as follows. Section \ref{Sys_model} briefly describes the problem formulation including the model of UAV's integrated guidance and control system. Section \ref{controller} presents the design of the proposed adaptive twisting sliding mode algorithm. Simulations and comparisons are introduced in Section \ref{results}. The paper ends with a conclusion and recommendation for future work.

\section{System model} \label{Sys_model}
A nonlinear dynamics of guidance and control combination for the UAV in the pitch plane is modeled in this section. 



\subsection{Geometric and nonlinear dynamics}
\begin{figure}
	\centering
	\includegraphics[width=8cm]{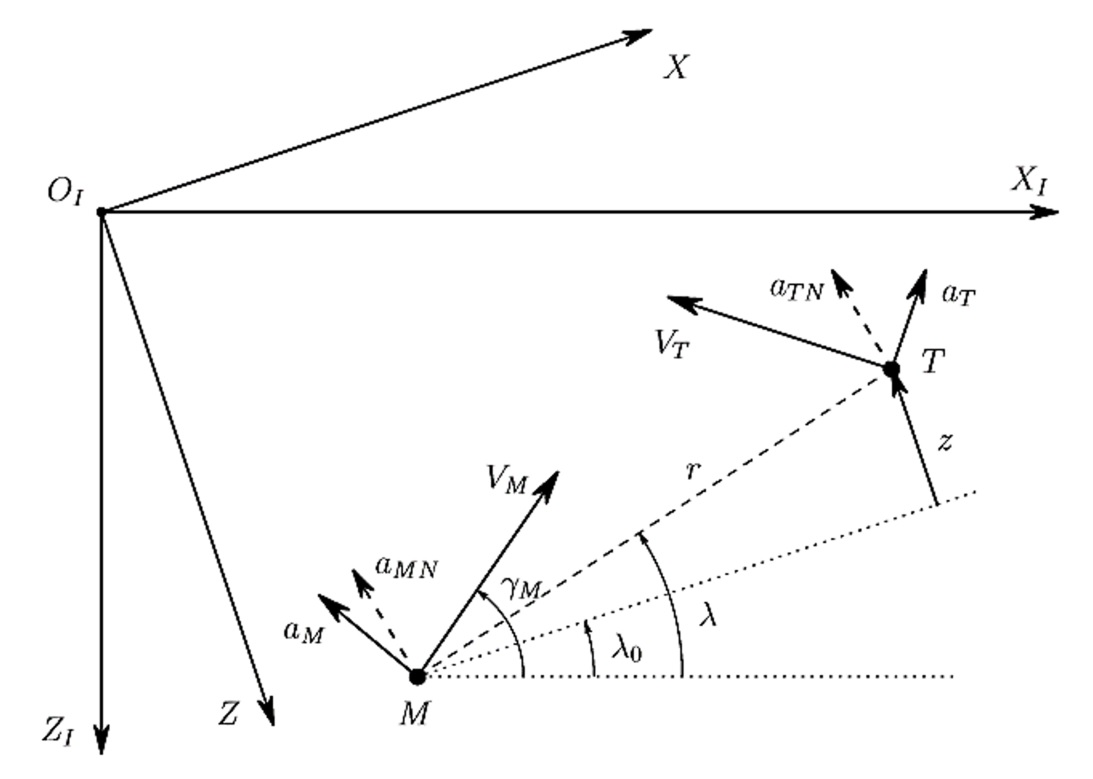}
	\caption{Correlation between the UAV and the target}
	\label{UAV_Target}
	\vspace{-3mm}
\end{figure}
The model of the UAV-target correlation used in this work is illustrated in Fig. \ref{UAV_Target}. Wherein, the inertial frame, $X_1 O_1 Z_1$, is defined by the ground with the $Z_1$ axis being directed down to the earth's center. The symbols $M$ and $T$ represent the positions of the UAV and the target, respectively. The velocity, normal acceleration, and flight-path angle of the UAV and the target are $V, a$ and  $\gamma$, respectively. The distance between the UAV and its target is $r$, and the angle between the line of sight (LOS) and the initial direction of the UAV is $\lambda$. Ignoring the gravitational acceleration, the UAV-target geometrical dynamic relationship in the polar coordinate system is as follows:
\begin{align}  \label{eqUAV_tarGeo}
	&\dot{r} = V_r,\\
	&\dot{\lambda} = V_\lambda/r,
\end{align}

From the Fig. \ref{UAV_Target}, we find the relative velocity between the UAV and the target as:
\begin{equation} \label{eqRelative_velo}
V_r = -[V_M \cos (\gamma_M - \lambda) + V_T \cos (\gamma_T + \lambda)],
\end{equation}
and the velocity perpendicular to LOS is:
\begin{equation}  \label{eqPerLOS_velo}
V_\lambda = -V_M \sin (\gamma_M - \lambda) + V_T \sin (\gamma_T + \lambda).
\end{equation} 

The remaining flight time to the target is defined as:  
\begin{equation}   \label{eqtgo}
t_{go} = -r/V_r.
\end{equation} 
\subsection{UAV and Target Dynamical Model}
The target movement velocity is supposed to be unchanged. Its dynamics can be represented as the following system of equations:
\begin{align}  \label{eqUAVtargetDyn}
	\begin{cases}
		&\dot{a}_T = (a^c_T - a_T)/\tau_T,\\
		&\dot{\gamma}_T = a_T/V_T,
	\end{cases}
\end{align}
where $\tau_T$ and $a^c_T$ are respectively the dynamics' time constant and acceleration command of the target.

For the UAV, by ignoring the impact of the thrust and its speed changes insignificantly, the mathematical model of the kamikaze UAV can be described as follows:
\begin{align}  \label{eqUAVtargetDyn}
	\begin{cases}
		&\dot{\alpha} = q-L(\alpha, \delta)/(mV_M),\\
		&\dot{q} = M(\alpha, q, \delta)/I,\\
		&\dot{\theta} = q,\\
		&\dot{\delta} = (\delta^c - \delta)/\tau_s,
	\end{cases}
\end{align}
where $\alpha$ is the attack angle, $q$ is the pitch angular velocity, $m$ is the mass and $M$ is the pitch moment of the UAV, $\delta$  is the angular rotation of the canard controlled by the actuator with time constant $\tau_s$, $T$ is the engine thrust along the longitudinal axis of the UAV; $L$ and $D$ are the lift and drag forces. 


\subsection{UAV - Target Dynamical Model} \label{subsecUAV_Target_Mod}
For complexity reduction, the SMC sliding surfaces designed for the integrated system will be represented using simplified mathematical models \cite{zhurbal2011}. Assuming that the approaching speed $V_r$ and time  $t_f$ are unchanged. Time $t_{go}$ from Eq. (\ref{eqtgo}) becomes $t_{go} = t_f - t$. $z$ is perpendicular to the initial LOS and represents the relative deviation between the target and the UAV.

The overall state system has the form:
\begin{equation}\label{eqOvr}
\dot{\textbf{x}} = A\textbf{x} + Bu + Cv,
\end{equation}

The Eq. (\ref{eqOvr}) can be rewritten for the UAV as follows:
\begin{equation}\label{eqUAVDyn}
\dot{\textbf{x}}_G = A_G\textbf{x}_G + B_G a^c_{MN} + G_G a^c_{TN},
\end{equation}
In which, $a_{TN}$ and $a_{MN}$ are the target and UAV's accelerations perpendicular to the LOS, respectively; $B_G = \begin{bmatrix} 0 & 0 & 0 & 1/\tau_M \end{bmatrix}^T$, $G_G = \begin{bmatrix} 0 & 0 & 1/\tau_T & 0 \end{bmatrix}^T$; with
\begin{equation}\label{eqAG}
A_G = \begin{bmatrix} A_{G11} & A_{G12} \\ [0]_{1 \times 3} & -1/\tau_M \end{bmatrix}, 
\end{equation}
$A_{G11} = \begin{bmatrix} 0 & 1 & 0 \\ 0 & 0 & 1\\ 0 & 0 & -1/\tau_T \end{bmatrix}$,
$A_{G12} = \begin{bmatrix} 0 \\ -1 \\ 0 \end{bmatrix}$, 
we have linearized state vector $\textbf{x}_G$ is defined by:
\begin{equation}\label{eqxG}
\textbf{x}_G = \begin{bmatrix} z & \dot{z} & a_{TN} & a_{MN}  \end{bmatrix}^T;
\end{equation}

 
\subsection{UAV Dynamical Model}
The UAV kinematics in a vertical plane is described in the system (\ref{eqUAVModel}):
\begin{equation}\label{eqUAVModel}
\dot{\textbf{x}}_M = A_M\textbf{x}_M + B_M \delta^c
\end{equation}
where, $\textbf{x}_M = \begin{bmatrix} \alpha & q & \delta \end{bmatrix}^T$ is the state vector of the UAV, $B_M = \begin{bmatrix} 0 & 0 & 1/\tau_s \end{bmatrix}^T$;
\begin{equation}\label{eqAM}
A_M = \begin{bmatrix} -L_\alpha/V_M & 1 & -L_\beta/V_M \\ M_\alpha & M_q & M_\delta\\ 0 & 0 & -1/\tau_s \end{bmatrix};
\end{equation}

The tangential acceleration of the UAV in Eq. (\ref{eqxG}) is the  control signal and the UAV and can be determined as:
\begin{equation}\label{eqAMN}
a_{MN} = C_M \dot{\textbf{x}}_M
\end{equation}

In which, $C_M = \begin{bmatrix} L_\alpha & 0 & L_\delta \end{bmatrix} \cos(\gamma_{M0} - \lambda_0)$

\subsection{Integrated Dynamical Model}
The integrated control can be generated from the overall model in Eq. (\ref{eqOvr}) as:
\begin{equation}\label{eqIntDyn}
\dot{\textbf{x}}_I = A_I\textbf{x}_I + B_I \delta^c + G_I a^c_{TN},
\end{equation}
in which, $\dot{\textbf{x}}_I = \begin{bmatrix} z & \dot{z} & a_{TN} & \alpha & q & \delta \end{bmatrix}^T$ is the state vector; $B_I = \begin{bmatrix} [0]_{1 \times 5} &  1/\tau_s \end{bmatrix}^T$; $G_I = \begin{bmatrix} 0 & 0 &  1/\tau_T & [0]_{1 \times 3} \end{bmatrix}^T$;\\
\begin{equation}\label{eqAGC}
A_I = \begin{bmatrix} A_{G11} & A_{12} \\ [0]_{3 \times 3} & A_M \end{bmatrix}, \text{ with } 
A_{12} =  \left[[0]_{1 \times 3} \text{ } -C_M \text{  } [0]_{1 \times 3}\right]^T
\end{equation}

Before proceeding further, a terminal projection strategy is presented in \cite{shaferman2015} for order decrease. This method converts the existing miss distance into Zero-Effort Miss distance (ZEM). The conversion for $\dot{\textbf{x}}$ in Eq. (\ref{eqIntDyn}) is:
\begin{equation}\label{eqZEM}
Z_I = C_I \Phi_I(t_{go})\textbf{x}_I
\end{equation}
where $C_I$ is a constant matrix that draws the fitting components of the state vector, $C_I = \begin{bmatrix} 1 & [0]_{1 \times 5} \end{bmatrix}$, $\Phi$ is the conversion matrix corresponding to Eq. (\ref{eqIntDyn}), i.e., \\$\Phi_I(t_{go}) = \exp(A_I t_{go})$. 

Assuming a small deviation in the impact triangle (Fig. \ref{UAV_Target}), $z$ is perpendicular to the initial LOS, the speed can be calculated as follows:
\begin{equation}\label{eqZ}
z \approx (\lambda - \lambda_0)r.
\end{equation}
Take the differential of Eq. (\ref{eqZ}), we will have
\begin{equation}\label{eqZdot}
z + \dot{z} t_{go}  = -V_r t_{go}^2 \dot{\lambda}.
\end{equation}

From Eqs. (\ref{eqtgo}), (\ref{eqZEM}), (\ref{eqZdot}) and (\ref{eqOvr}), we finally find the ZEM for the integrated system as:
\begin{equation}\label{eqfinalZGC}
Z_I = -V_r t_{go}^2 \dot{\lambda} + a_{TN} \tau^2_T \psi + C_I \Phi_I(t_{go})\mathbf{\bar{x}}_I,
\end{equation}
therein, 
\begin{equation}\label{eqbarX}
\mathbf{\bar{x}}_I = \begin{bmatrix} [0]_{1 \times 3} & \alpha & q &\delta \end{bmatrix}^T,
\end{equation}
with
\begin{equation*}
\psi = e^{-\frac{t_{go}}{\tau_T}} + \frac{t_{go}}{\tau_T} - 1, 
\end{equation*}
\begin{equation}\label{eqdotAlpBet}
\dot{\alpha} = q -  \frac{\bar{a}_M + \Delta a}{V_M}, \dot{\delta} = \frac{\delta^c - \delta}{\tau_s}, 
\end{equation}
\begin{equation}\label{eqdotq}
\dot{q} = \bar{M}^B_\alpha \bar{f}_3(\alpha) + \bar{M}_q q + \bar{M}_\delta \bar{f}_4(\alpha + \delta) + \Delta_q, 
\end{equation}
in which, $\bar{M}_{(\cdot)}$ and $\bar{f}_i(\cdot), i = 1, \cdots, 4$ are some approximations of $(\cdot)$, $\Delta_q$ is the highest value of the pitching moment, i.e., $\abs{\Delta_q} \leq \bar{\Delta}_q$.

\section{Control Design} \label{controller}
Given the desired state vector $\dot{\textbf{x}}_I = \begin{bmatrix} z & \dot{z} & a_{TN} & \alpha & q & \delta \end{bmatrix}^T$ in Eq. (\ref{eqIntDyn}), the overall control law is proposed as:
\begin{equation}\label{eqUoverall}
u(t)= u_{eq}(t) + u_D(t),
\end{equation}
where $u_{eq}(t)$ and $u_D(t)$  are respectively the equivalent control and the discontinuous part containing switching elements. 

For the integrated control system, the input is the elevator command $\delta_c$, and the acceleration command $a_{TN}^c$ of the target is considered a disturbance. The ZEM in Eq. (\ref{eqfinalZGC}), which is desired to drive to zero, will be regarded as the sliding surface: 
\begin{equation}\label{Eq10}
\mathbf{\sigma} = Z_I,
\end{equation}

\subsection{Design $u_{eq}$}
Taking the time derivative of $\sigma$, we have:
\begin{align} \label{Eq14a}
\dot{\sigma} = &-V_r t_{go}^2 \ddot{\lambda} - \dot{V} t_{go}^2 \dot{\lambda} - 2 V_r t_{go} \dot{t}_{go} \dot{\lambda} + \dot{a}_{TN} \tau^2_T \psi +  \\ \nonumber
				&+ a_{TN} \tau^2_T \psi' \dot{t}_{go} + C_I \Phi'_I(t_{go})\bar{\textbf{x}}_I \dot{t}_{go} + C_I \Phi_I(t_{go})\dot{\bar{\textbf{x}}}_I.
\end{align}
where
\begin{equation}
\psi' = \frac{\partial \psi}{\partial t_{go}} = \frac{t_{go}/\tau_T - \psi}{\tau_T}
\end{equation}
or
\begin{align}\label{eqdotSigma}\nonumber
\dot{\sigma} = &\left[ V_\lambda + a_{TN} \tau_T \left(1 - e^{-\frac{t_{go}}{\tau_T}}\right)\right] \dfrac{\dot{V}_r r}{V^2_r} 
				+ \tau_T\left( a_{TN}^c + \tau_T \Delta a_{TN} \right) \psi \\ 
				&- t_{go} a_{MN} + C_I \left[ \Phi'_I(t_{go})  \bar{\textbf{x}}_I \dot{t}_{go}  + \Phi_I(t_{go})\dot{\bar{\textbf{x}}}_I\right].
\end{align}
in Eq. (\ref{eqdotSigma}), $\ddot{\lambda}, \dot{V}_r$, and $\dot{V}_\lambda$ are given as:
\begin{align}
\ddot{\lambda} &= \frac{\dot{V}_\lambda}{r} - \frac{V_\lambda V_r}{r^2}, \\
\dot{V}_r &= \frac{V^2_\lambda}{r} + a_M \sin(\gamma_M - \lambda) + a_T \sin(\gamma_T + \lambda), \\
\dot{V}_\lambda &= -\frac{V_\lambda V_r}{r} - a_M \cos(\gamma_M - \lambda) + a_T \cos(\gamma_T + \lambda),
\end{align}
$t_{go}$ can be found in according with the following relation:
\begin{equation}\label{eqdottGo}
\dot{t}_{go} = -1 +\dfrac{\dot{V}_r r}{V^2_r},
\end{equation}
and 
\begin{equation}\label{eqdevPhi}
\Phi'_I(t_{go}) = \dfrac{\partial \Phi_I(t_{go})}{\partial t_{go}} = A_I \Phi_I(t_{go}).
\end{equation}

By setting $\bar{\textbf{y}}_I = A_I \bar{\textbf{x}}_I, \bar{a}_{MN} = - \bar{\textbf{y}}_I (2)$, and $\Phi_I^{(1,6)}(t_{go})$ are the vector contains the first and sixth elements of $\Phi_I(t_{go})$, then substituting Eqs. (\ref{eqdotAlpBet}), (\ref{eqdottGo}), (\ref{eqdotq}) and (\ref{eqdevPhi}) to (\ref{eqdotSigma}), we found the overall sliding manipulator as: 
\begin{align}\label{eqSlidMani}\nonumber
\dot{\sigma} = &\left[ V_\lambda + a_{TN} \tau_T \left(1 - e^{-\frac{t_{go}}{\tau_T}}\right) + C_I \Phi_I(t_{go})\bar{\textbf{y}}_I \right] \dfrac{\dot{V}_r r}{V^2_r} \\
				& + \Phi_I^{(1,6)}(t_{go}) \frac{\delta^c}{\tau_s} + \tau_T (a^c_{TN} + \tau_T \Delta a_{TN})\psi + \Delta_I.
\end{align}

Once the sliding mode is generated, the control variable $u$ can be regarded as the equivalent control $u_{eq}$. To confirm the derivative of the sliding surface equals zero, we can derive the equivalent control law as follows:
\begin{align}\label{eqUeq}
u_{eq} = &-\left[ V_\lambda + a_{TN} \tau_T \left(1 - e^{-\frac{t_{go}}{\tau_T}}\right) + C_I \Phi_I(t_{go})\bar{\textbf{y}}_I \right] \dfrac{\dot{V}_r r \tau_s}{V^2_r \Phi_I^{(1,6)}(t_{go})}
\end{align}

\subsection{Design $u_D$}
The twisting controllers is selected for the discontinuous part of $u$ in Eq. (\ref{eqUoverall}) as:
\begin{align}\label{eqUtwist}
u_T = \begin{cases} 
-\mu \beta \text{sign}(\sigma) & \text{if} \enskip \sigma\dot{\sigma} \leq 0 \\
-\beta \text{sign}(\sigma) & \text{if} \enskip \sigma\dot{\sigma} > 0,
\end{cases}
\end{align}
where $\mu < 1$ is a fixed positive number and $\beta > 0$ is the control gain.

To improve the transient tracking and control performance, $\beta$ can be decided to meet the one-stage accelerated convolutional algorithm condition  \cite{Dvir2015}, i.e., $\beta_i = \text{max} \{ \beta_*, \gamma \abs{\sigma}^{\rho}\}$, with $\beta_* > 0$, $\gamma > 0$ and $\rho > 0$ are some constants. Besides that, due to the need for constant time stability over a large UAV operating area, we advocate for an adaptive adjustment of the gain $\beta_i$ in (\ref{eqUtwist}) following the innovative approach outlined in \cite{hoang2017}. This adjustment will be developed based on the below equation:
\begin{align}\label{Eq24}
\dot{\beta} &= \begin{cases}
\bar{\omega} \abs{\sigma(\omega,t)} \text{sign}(|\sigma(\omega,t)|^{\rho}-\epsilon) &\text{if} \enskip \beta > \beta_m\\
\eta &\text{if} \enskip\beta \leq \beta_m,
\end{cases}
\end{align}
where $\bar{\omega} > 0, \rho > 0$, $\epsilon> $, and $\eta> 0$ are constants, $\beta_m$ is an adaptation threshold, such as $\beta_m  > \beta_*$.

In trying to find a condition for the convergence of the proposed control and adaptation schemes, let us consider the Lyapunov function candidate:

\begin{equation}\label{eqV}
V = \frac{1}{2} \sigma^2 + \dfrac{1}{2\gamma}(\beta -\beta_M)^2,
\end{equation}
where $\gamma$ represents a positive constant, and $\beta_M$ denotes the maximum value of the adaptive gain, specifically satisfying $0 < \beta_m < \beta < \beta_M$. When we take the time derivative of $V$ and substitute $\dot{\sigma}$ from (\ref{eqSlidMani}), we obtain a result such as:
\begin{align}\label{eqVdot1}
\dot{V} =& \sigma\dot{\sigma} + \dfrac{1}{\gamma}(\beta -\beta_M)\dot{\beta}. 
\end{align}

Motivated by \cite{shima2006, zhurbal2011} and \cite{hoang2017}, Eq. (\ref{eqVdot1}) can be rewritten for the case $\sigma\dot{\sigma} \leq 0$ as,
\begin{align}\label{eqVdot2}
\dot{V} &= -\abs{\sigma_I}(\mu_I - \bar{\Delta}a_{TNc} - \bar{\Delta}a_{TN\tau} - \bar{\Delta}a_I) \nonumber \\
&+\dfrac{\bar{\omega}}{\gamma\mu}(\beta -\beta_M)  \text{sign}(|\sigma|^{\rho}-\epsilon)
\end{align}

By assuming that both the true dynamics of the UAV ($\Delta a_{MN}$), the target ($\Delta a_{TN}$), and the integrated guidance and control modeling errors are limited by $\bar{\Delta}a_{MN}, \bar{\Delta}a_{TN}, \bar{\Delta}_I$, respectively; also, the last three terms in Eq. \ref{eqSlidMani} are bounded, such that:
$\abs{\tau_T a^c_{TN} \psi} \leq \bar{\Delta}a_{TNc}$,
$ \abs{\tau_T a_{TN} \psi} \leq \bar{\Delta}a_{TN\tau}$,
$ \abs{\Delta_I} \leq \bar{\Delta}_I$.
Besides that, with sufficiently small $\epsilon$ such that $|\sigma|^{\rho}>\epsilon$ \cite{kumar2015}, $\dot{V} \leq 0$ if
\begin{equation}
\mu_I >  \bar{\Delta}a_{TNc} + \bar{\Delta}a_{TN\tau} + \bar{\Delta}_I.
\end{equation}
Noting that only the case $\alpha >\alpha_M$ is considered here as otherwise the last term in the right hand side of (\ref{eqVdot2}) becomes $\dfrac{1}{\gamma}(\alpha -\alpha_M)\eta < 0$. 

For the case $\sigma\dot{\sigma} > 0$, from (\ref{eqUtwist}), we can hold the identical consequence as above if assuming  $\mu$ is equal to one.

\section{Results and discussion} \label{results}
\subsection{Simulation setup} 
\begin{figure}
	\centering
	\includegraphics[width=7cm,clip]{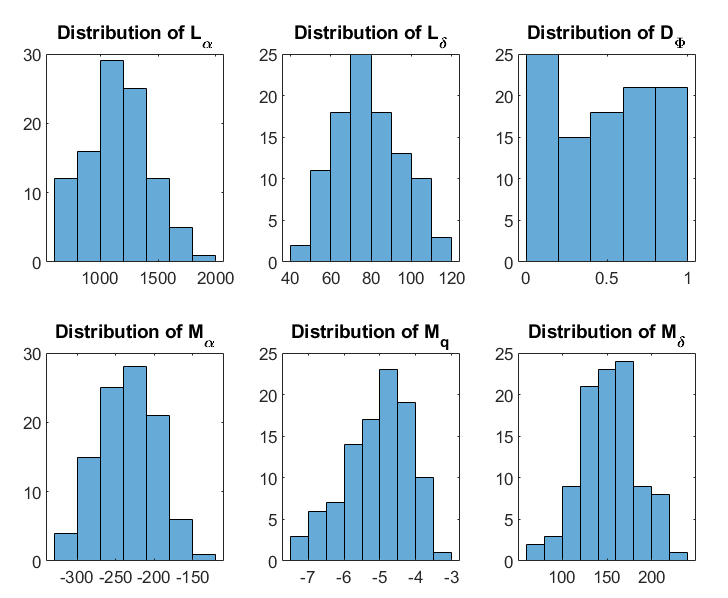}
	\caption{System parameters with modeling errors}
	\label{Histogram}
\end{figure}

\begin{figure}
	\centering
	\includegraphics[width=7cm,clip]{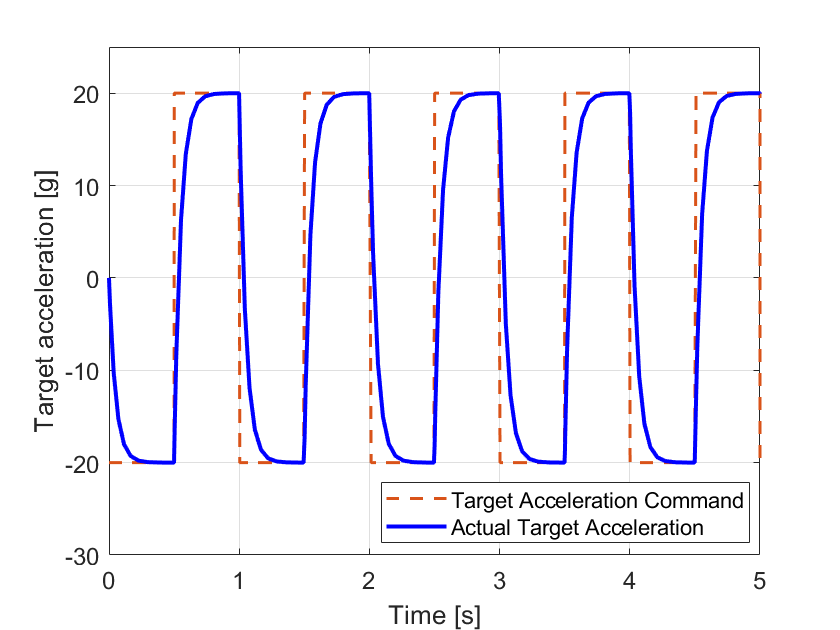}
	\caption{Target maneuver}
	\label{Target_move}
\end{figure}

In the simulation, the task for the UAV is to attack a target in the distance of 3000 meters with the integrated guidance and control algorithm using ATSMC. The initial positions and states of both UAV and its target are taken from \cite{shima2006}.  In which the UAV parameters are $V_M = 380 \text{m/s}$, $\bar{L}^B_\alpha = 1190 \text{m/s}^2$, $\bar{L}_\delta = 80 \text{m/s}^2$, $\bar{M}^B_\alpha = -234 \text{s}^{-2}$, $\bar{M}_q = -5 \text{s}^{-1}$, $\bar{M}_\delta = 160 \text{s}^{-2}$, $a^{max}_M = 40 \text{g}$, and $\tau_M = 0.1 \text{s}$. Its carnard servo time constant is $\tau_s = 0.02 \text{s}$. The target parameters are $V_T = 380 \text{m/s}$, $a^{max}_T = 20 \text{g}$, $\Delta T = 1 \text{s}$, $\Delta \phi \in [0, 1] \text{s}$, and $\tau_T \in [0.05, 0.2] \text{s}$. In this work, the terms $\bar{L}^B_\alpha $, $\bar{L}_\delta$, $\bar{M}^B_\alpha$, $\bar{M}_q$, $\bar{M}_\delta$  have been added uncertainties, which is contributed to 20$\%$ as a normally distributed function, as shown in Fig. \ref{Histogram}. The target maneuver is entirely executed in a square wave acceleration with a period $\Delta T$ and a phase $\Delta \phi$ (Fig. \ref{Target_move})  to evade attack by the UAV.

The variables and their corresponding values of the controller used in this work are given in Table \ref{Table2}. The performance of our control design is demonstrated in simulation and comparison with the conventional sliding control (SMC) and the twisting sliding mode (TSMC).
\begin{table}[h] 
\centering
	\caption{Control parameters} \label{Table2}
	\begin{tabular}{|l|l|l|l|}
		\hline
		Variable 	& Value 			& Variable 	& Value \\ \hline
		$\gamma$	 	& 0.25	 		& $\bar{\omega}$	& 80.65\\
		$\mu_I$ 		& 0.7	 		& $\beta_m$		& 0.01\\
		$\epsilon$	& 0.6			& $\beta_M$		& 1.57\\
		$\beta_0$	& 1.57	 		& $\eta$			& 0.05\\
		\hline 
	\end{tabular}
\end{table}

\begin{figure}
	\centering
	\includegraphics[width=8cm,clip]{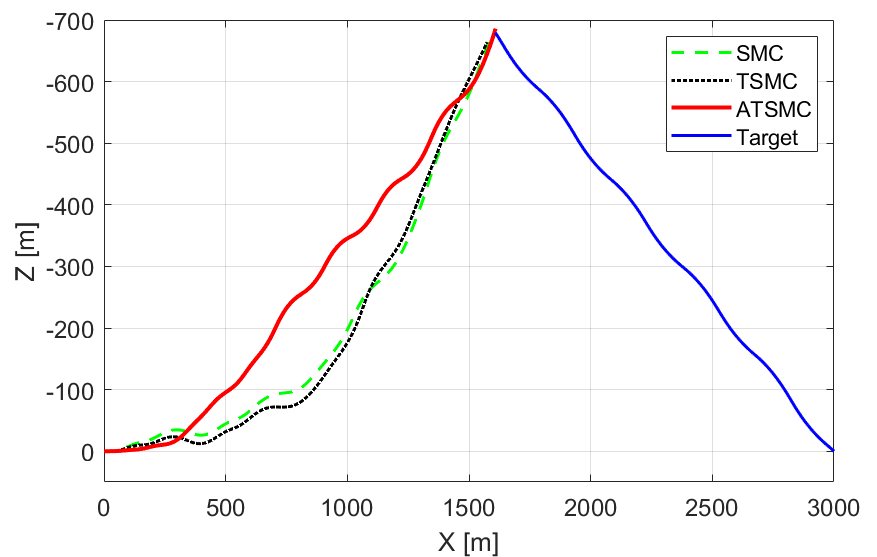}
	\caption{Engagement trajectories of the target and the UAV with distinctive controllers}
	\label{EngageTraj}
\end{figure}

The engagement scenarios together with the target trajectory, and the UAV trajectories plotted in different schemes are presented in Fig. \ref{EngageTraj}. Wherein, the UAV's initial velocity is adjusted in corresponding with the foremost Light of Sight (LOS). The frequent motion of the target and the accordingly adjusted movement of the UAV are clearly seen in the figure. The LOS direction stays about unchanged even though there is a heading error at the beginning. All three control schemes show almost equivalent performance. Among them, the ATSMC suggests a more stable, and better trajectory in terms of smoothness.

\begin{figure}
	\centering
	\includegraphics[width=7cm,clip]{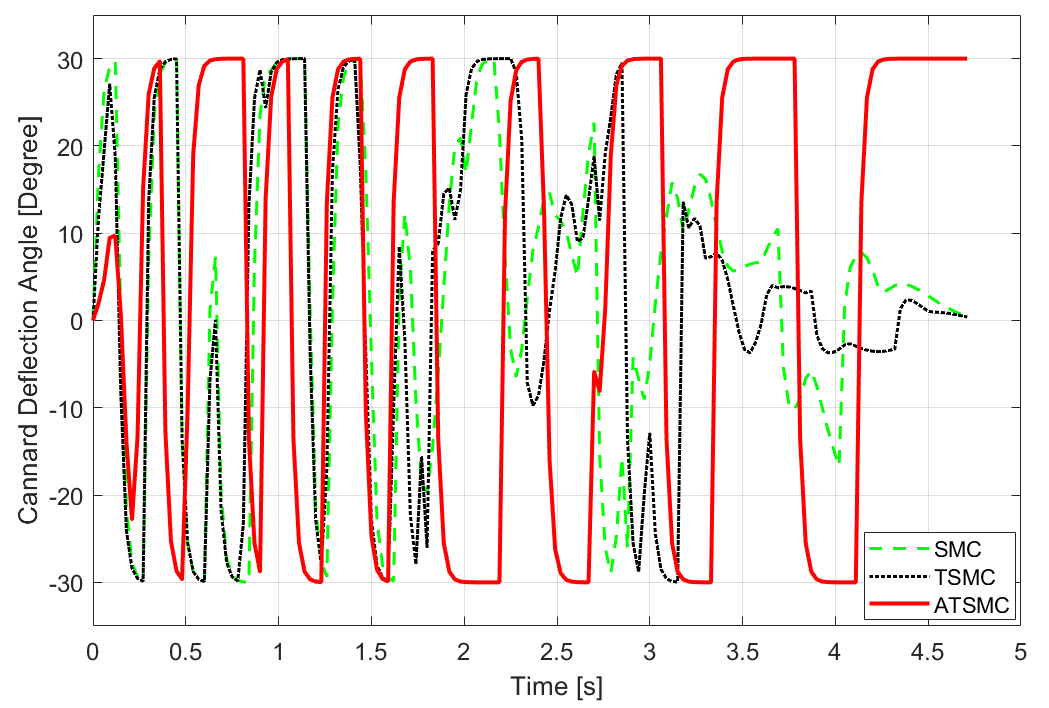}
	\caption{Cannar deflection of different SMC variations}
	\label{CannarDeflection}
\end{figure}

Fig. \ref{CannarDeflection} illustrates the canard deflection reaction, with its velocity being strict in $\dot{\delta} = [ -30,30]$ degrees per second and $\abs{\delta} \leq 30^0$ during flight. The very oscillatory during the flight of saturated canard deflections are marked with the ATSMC controllers as the quick adaptition to preserve the UAV on the shortest trajectory (Fig. \ref{EngageTraj}).

\begin{figure}
	\centering
	\includegraphics[width=8cm,clip]{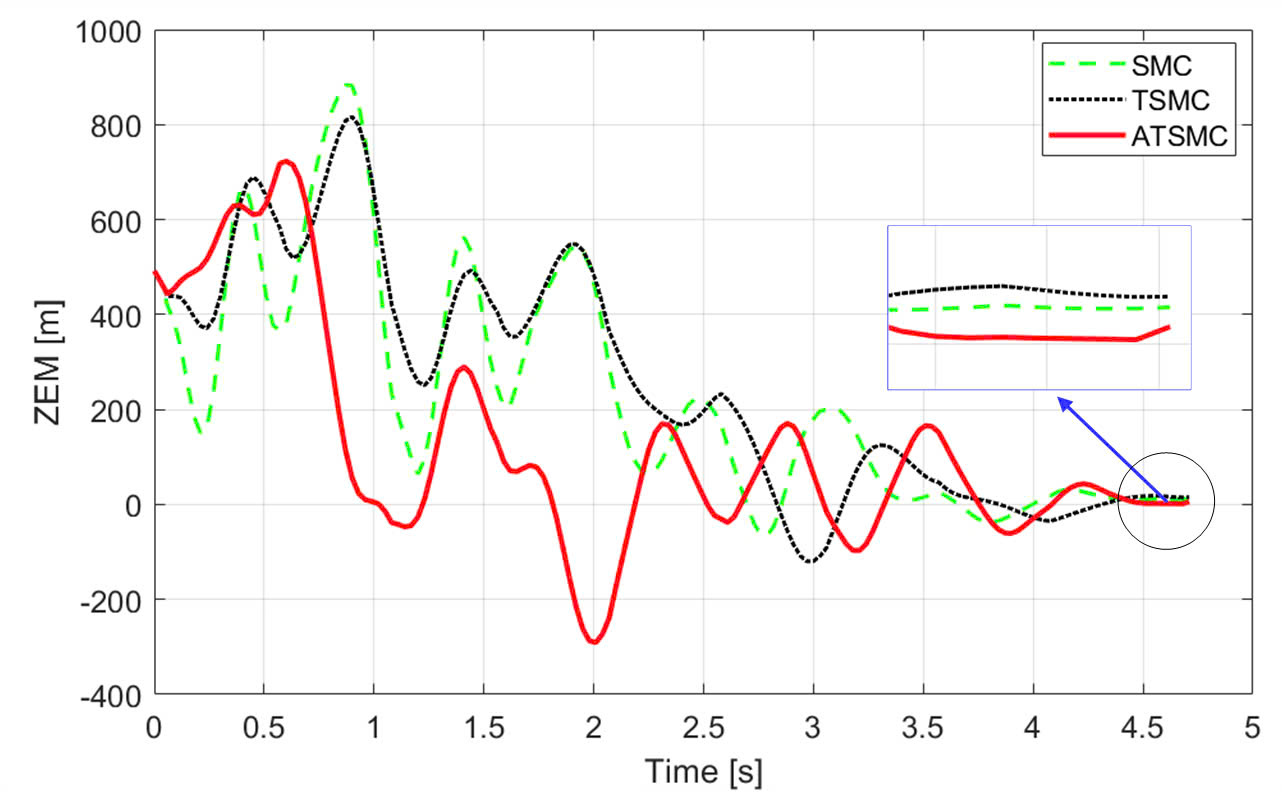}
	\caption{Comparison of Zero Effort Miss Distances}
	\label{ZeroEffortMiss}
\end{figure}

The simulation results for ZEM are presented in Fig. \ref{ZeroEffortMiss} with the last terminal phase zoomed in. The figure shows that even large initial distance (3000 m) and somewhat abrupt variations of the target trajectory (characterized by a continuously small time constant), the three algorithms are capable of guiding the integrated system reach to the optimum miss. Contrary to the SMC and TSMC, our proposed method exhibits a lesser overshoot in ZEM and the smallest value at the engagement phase. 

\begin{figure}
	\centering
	\includegraphics[width=8cm]{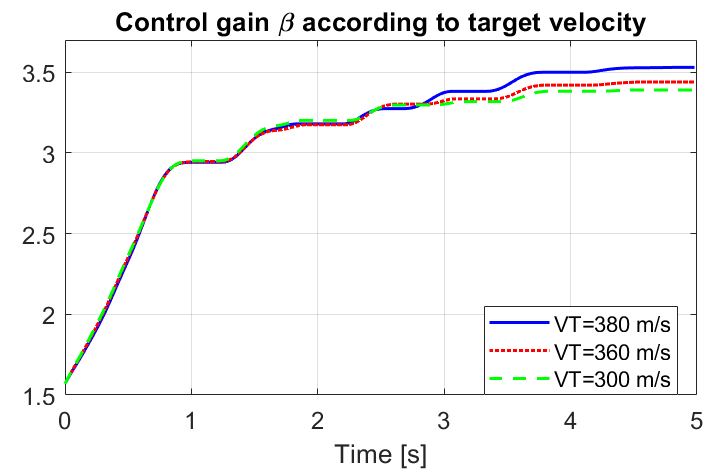}
	\caption{Variation of adaptation gains}
	\label{ControlGain}
\end{figure}

The deviation of the adaptive gain found in the simulation is displayed in Fig. \ref{ControlGain} in three different velocities of the target, i.e., $v_T = 300, 360, 380$ m/s. More elevated gain magnitudes indicate more power is demanded to stabilize the integrated system. This result, hence, demonstrates the validity of our adaptation scheme.

\section{Conclusion}
In this paper, we have proposed an adaptive technique of twisting sliding mode for an integrated vertical UAV autopilot and guidance system when attacking a target. The significance of the algorithm is motivated by the accelerated twisting sliding mode control to an adaptive scheme. We derive the sliding manifold employing the zero-effort miss distance. The effectiveness of the ATSMC is evaluated and compared with two other SMC-based methods in a number of simulation scenarios. Results indicate that the integrated design has the capability of enhancing interception accuracy against strong disturbances, nonlinearity, and uncertainties; especially, tremendous variation in target trajectory and its speed. Our future work will focus on implementing the proposed controller to enable higher-level tasks of the UAV such as cooperatively working in a swarm or formation.

%
%
%

%
%
\bibliography{IEEEabrv,bibi}
\end{document}